\documentclass[sigconf]{acmart}
\renewcommand\footnotetextcopyrightpermission[1]{} 
\usepackage{booktabs} 

\usepackage[english]{babel}
\usepackage[utf8x]{inputenc}
\usepackage{amsmath, bm}
\usepackage{amsthm}
\usepackage{amsfonts}
\usepackage[colorinlistoftodos]{todonotes}
\usepackage{booktabs}
\usepackage{multirow}
\usepackage{enumitem}
\usepackage{mleftright}
\usepackage{diagbox}

\newcommand{\vecx}{{\bf x}}
\newcommand{\vecy}{{\bf y}}
\newcommand{\vecw}{{\bf w}}
\newcommand{\vecb}{{\bf b}}
\newcommand{\vech}{{\bf h}}
\newcommand{\vecv}{{\bf v}}
\newcommand{\veci}{{\bf i}}
\newcommand{\vecj}{{\bf j}}
\newcommand{\vecalpha}{{\bm \alpha}}

\newcommand{\veczero}{{\bf 0}}

\addto\extrasenglish{%
}

\definecolor{cornellred}{rgb}{0.7, 0.11, 0.11}

\begin{document}
\title{Deep \& Cross Network for Ad Click Predictions}

\author{Ruoxi Wang}
\affiliation{%
  \institution{Stanford University}
  \city{Stanford} 
  \state{CA} 
}
\email{ruoxi@stanford.edu}

\author{Bin Fu}
\affiliation{%
  \institution{Google Inc.}
  \city{New York} 
  \country{NY}}
\email{binfu@google.com}

\author{Gang Fu}
\affiliation{%
  \institution{Google Inc.}
  \city{New York} 
  \state{NY} 
	}
\email{thomasfu@google.com}

\author{Mingliang Wang}
\affiliation{
  \institution{Google Inc.}
  \city{New York} 
  \country{NY}}
\email{mlwang@google.com}

\renewcommand{\shortauthors}{R. Wang et al.}

\begin{abstract}
Feature engineering has been the key to the success of many prediction models. However, the process is nontrivial and often requires manual feature engineering or exhaustive searching. {DNN}s are able to automatically learn feature interactions; however, they generate all the interactions implicitly, and are not necessarily efficient in learning all types of cross features.  In this paper, we propose the Deep \& Cross Network (DCN) which keeps the benefits of a {DNN} model, and beyond that, it introduces a novel cross network that is more efficient in learning certain bounded-degree feature interactions. In particular, {DCN} explicitly applies feature crossing at each layer, requires no manual feature engineering, and adds negligible extra complexity to the {DNN} model. Our experimental results have demonstrated its superiority over the state-of-art algorithms on the CTR prediction dataset and dense classification dataset, in terms of both model accuracy and memory usage.
\end{abstract}

%
%


\maketitle

\section{Introduction}
Click-through rate (CTR) prediction is a large-scale problem that is essential to multi-billion dollar online advertising industry. In the advertising industry, advertisers pay publishers to display their ads on publishers' sites. One popular payment model is the cost-per-click (CPC) model, where advertisers are charged only when a click occurs. As a consequence, a publisher's revenue relies heavily on the ability to predict {CTR} accurately.

Identifying frequently predictive features and at the same time exploring unseen or rare cross features is the key to making good predictions. However, data for Web-scale recommender systems is mostly discrete and categorical, leading to a large and sparse feature space that is challenging for feature exploration. This has limited most large-scale systems to linear models such as logistic regression.  

Linear models \cite{chapelle2015simple} are simple, interpretable and easy to scale; however, they are limited in their expressive power. Cross features, on the other hand, have been shown to be significant in improving the models' expressiveness. Unfortunately, it often requires manual feature engineering or exhaustive search to identify such features; moreover, generalizing to unseen feature interactions is difficult. 

In this paper, we aim to avoid task-specific feature engineering by introducing a novel neural network structure -- a \emph{cross network} -- that explicitly applies feature crossing in an automatic fashion. The cross network consists of multiple layers, where the highest-degree of interactions are provably determined by layer depth. Each layer produces higher-order interactions based on existing ones, and keeps the interactions from previous layers. 
We train the cross network jointly with a deep neural network (DNN) \cite{lecun2015deep,schmidhuber2015deep}. {DNN} has the promise to capture very complex interactions across features; however, compared to our cross network it requires nearly an order of magnitude more parameters, is unable to form cross features explicitly, and may fail to efficiently learn some types of feature interactions. Jointly training the cross and DNN components together, however, efficiently captures predictive feature interactions, and delivers state-of-the-art performance on the Criteo CTR dataset.

\subsection{Related Work}
Due to the dramatic increase in size and dimensionality of datasets, a number of methods have been proposed to avoid extensive task-specific feature engineering, mostly based on embedding techniques and neural networks.

Factorization machines (FMs) \cite{rendle2010factorization, rendle:tist2012} project sparse features onto low-dimensional dense vectors and learn feature interactions from vector inner products. Field-aware factorization machines (FFMs) \cite{juan2016field, juan2017field} further allow each feature to learn several vectors where each vector is associated with a field. Regrettably, the shallow structures of {FMs} and {FFMs} limit their representative power. There have been work extending FMs to higher orders \cite{blondel2016higher, yang2015tensor}, but one downside lies in their large number of parameters which yields undesirable computational cost.  Deep neural networks (DNN) are able to learn non-trivial high-degree feature interactions due to embedding vectors and nonlinear activation functions. The recent success of the Residual Network \cite{he2015deep} has enabled training of very deep networks. Deep Crossing \cite{shan2016deep} extends residual networks and achieves automatic feature learning by stacking all types of inputs. 

The remarkable success of deep learning has elicited theoretical analyses on its representative power. There has been research \cite{valiant2014learning, NIPS2016_6556} showing that {DNN}s are able to approximate an arbitrary function under certain smoothness assumptions to an arbitrary accuracy, given sufficiently many hidden units or hidden layers. Moreover, in practice, it has been found that {DNN}s work well with a feasible number of parameters. One key reason is that most functions of practical interest are not arbitrary.

Yet one remaining question is whether {DNN}s are indeed the most efficient ones in representing such functions of practical interest.  In the Kaggle\footnote{https://www.kaggle.com/} competition, the manually crafted features in many winning solutions are low-degree, in an explicit format and effective. The features learned by {DNN}s, on the other hand, are implicit and highly nonlinear.  This has shed light on designing a model that is able to learn bounded-degree feature interactions more efficiently and explicitly than a universal {DNN}.

The wide-and-deep \cite{cheng2016wide} is a model in this spirit. It takes cross features as inputs to a linear model, and jointly trains the linear model with a {DNN} model. However, the success of wide-and-deep hinges on a proper choice of cross features, an exponential problem for which there is yet no clear efficient method.  

\subsection{Main Contributions}
In this paper, we propose the Deep \& Cross Network ({DCN}) model that enables Web-scale automatic feature learning with both sparse and dense inputs. {DCN} efficiently captures effective feature interactions of bounded degrees, learns highly nonlinear interactions, requires no manual feature engineering or exhaustive searching, and has low computational cost.

The main contributions of the paper include:
\begin{itemize}[leftmargin=*]
	\item We propose a novel cross network that explicitly applies feature crossing at each layer, efficiently learns predictive cross features of bounded degrees, and requires no manual feature engineering or exhaustive searching. 
    \item The cross network is simple yet effective. By design, the highest polynomial degree increases at each layer and is determined by layer depth. The network consists of all the cross terms of degree up to the highest, with their coefficients all different. 
    \item The cross network is memory efficient, and easy to implement. 
    \item Our experimental results have demonstrated that with a cross network, {DCN} has lower logloss than a {DNN} with nearly an order of magnitude fewer number of parameters. 
\end{itemize}

The paper is organized as follows: \autoref{sec:deep_cross_network} describes the architecture of the Deep \& Cross Network. \autoref{sec:analysis_cross_network} analyzes the cross network in detail. \autoref{sec:experimental_results} shows the experimental results.


\section{Deep \& Cross Network ({DCN})}
\label{sec:deep_cross_network}
\begin{figure}[htbp]
  \centering
  \includegraphics[width=3.4in]{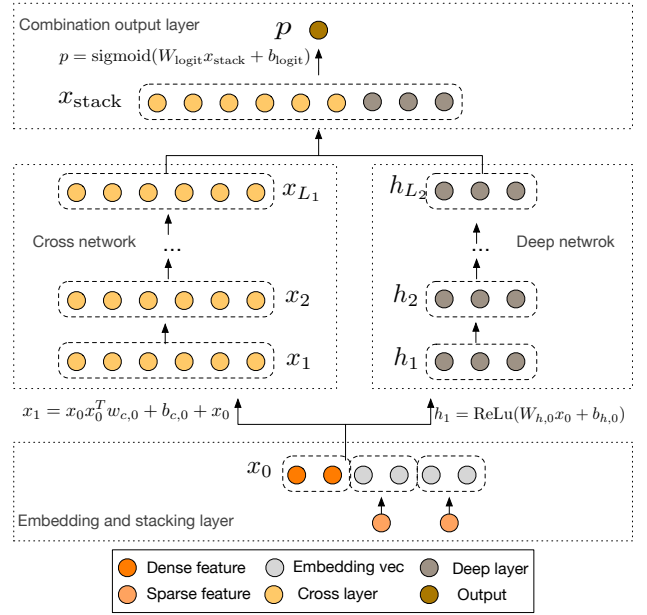}
  \caption{The Deep \& Cross Network}
  \label{fig:deep_cross_network}
\end{figure}

In this section we describe the architecture of Deep \& Cross Network {(DCN)} models.
A {DCN} model starts with an \emph{embedding and stacking layer}, followed by a \emph{cross network} and a \emph{deep network} in parallel. These in turn are followed by a final \emph{combination layer} which combines the outputs from the two networks. The complete {DCN} model is depicted in \autoref{fig:deep_cross_network}. 

\subsection{Embedding and Stacking Layer}
\label{sec:embedding}
We consider input data with sparse and dense features. In Web-scale recommender systems such as CTR prediction, the inputs are mostly categorical features, \emph{e.g.} {\ttfamily{"country=usa"}}. Such features are often encoded as one-hot vectors \emph{e.g.} {\ttfamily{"[0,1,0]"}}; however, this often leads to excessively high-dimensional feature spaces for large vocabularies.  

To reduce the dimensionality, we employ an embedding procedure to transform these binary features into dense vectors of real values (commonly called embedding vectors):
\begin{equation}
	\vecx_{\text{embed}, i} = W_{\text{embed}, i}\vecx_i,
\end{equation}
where $\vecx_{\text{embed}, i}$ is the embedding vector, $\vecx_i$ is the binary input in the $i$-th category, and $W_{\text{embed}, i} \in \mathbb{R}^{n_e \times n_v}$ is the corresponding embedding matrix that will be optimized together with other parameters in the network, and $n_e, n_v$ are the embedding size and vocabulary size, respectively.

In the end, we stack the embedding vectors, along with the normalized dense features $\vecx_{\text{dense}}$, into one vector:
\begin{equation}
	\vecx_0 = \left[\vecx_{\text{embed}, 1}^T, \ldots,\vecx_{\text{embed}, k}^T, \vecx_{\text{dense}}^T\right],
\end{equation}
and feed $\vecx_0$ to the network.

\subsection{Cross Network}
\label{sec:cross}
\begin{figure}[htbp]
  \centering
  \includegraphics[width=3in]{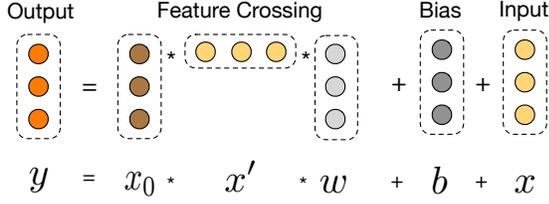}
  \caption{Visualization of a cross layer.}
  \label{fig:cross_sub_network}
\end{figure}

The key idea of our novel cross network is to apply explicit feature crossing in an efficient way. The cross network is composed of cross layers, with each layer having the following formula:

\begin{equation}
\label{eq:cross_layer}
	\vecx_{l+1} =\vecx_0\vecx_l^T \vecw_l + \vecb_l +\vecx_l = f(\vecx_l, \vecw_l, \vecb_l) +\vecx_l,
\end{equation}

where $\vecx_l,\vecx_{l+1} \in \mathbb{R}^{d}$ are column vectors denoting the outputs from the $l$-th and $(l+1)$-th cross layers, respectively; $\vecw_l, \vecb_l \in \mathbb{R}^{d}$ are the weight and bias parameters of the $l$-th layer. Each cross layer adds back its input after a feature crossing $f$, and the mapping function $f: \mathbb{R}^{d} \mapsto \mathbb{R}^{d}$ fits the residual of $\vecx_{l+1} -\vecx_l$. A visualization of one cross layer is shown in \autoref{fig:cross_sub_network}.

\textbf{High-degree Interaction Across Features.}
The special structure of the cross network causes the degree of cross features to grow with layer depth. The highest polynomial degree (in terms of input $\vecx_0$) for an $l$-layer cross network is $l+1$. In fact, the cross network comprises all the cross terms $x_1^{\alpha_1}x_2^{\alpha_2}\ldots x_d^{\alpha_d}$ of degree from 1 to $l+1$. Detailed analysis is in \autoref{sec:analysis_cross_network}.

\textbf{Complexity Analysis.}
Let $L_c$ denote the number of cross layers, and $d$ denote the input dimension. Then, the number of parameters involved in the cross network is
\begin{align*}
d \times L_c \times 2.
\end{align*} 
The time and space complexity of a cross network are linear in input dimension. 
Therefore, a cross network introduces negligible complexity compared to its deep counterpart, keeping the overall complexity for {DCN} at the same level as that of a traditional {DNN}. 
This efficiency benefits from the rank-one property of $\vecx_0\vecx_l^T$, which enables us to generate all cross terms without computing or storing the entire matrix.

The small number of parameters of the cross network has limited the model capacity. To capture highly nonlinear interactions, we introduce a deep network in parallel.

\subsection{Deep Network}
\label{sec:deep}
The deep network is a fully-connected feed-forward neural network, with each deep layer having the following formula:
\begin{equation}
	\vech_{l+1} = f(W_l\vech_l + \vecb_l),
\end{equation}
where $\vech_l \in \mathbb{R}^{n_l}, \vech_{l+1}\in \mathbb{R}^{n_{l+1}}$ are the $l$-th and $(l+1)$-th hidden layer, respectively; $W_l \in \mathbb{R}^{n_{l+1} \times n_l}, \vecb_l \in \mathbb{R}^{n_{l+1}}$ are parameters for the $l$-th deep layer; and $f(\cdot)$ is the ReLU function. 

\textbf{Complexity Analysis.}
For simplicity, we assume all the deep layers are of equal size. Let $L_d$ denote the number of deep layers and $m$ denote the deep layer size. Then, the number of parameters in the deep network is
\begin{align*}
d \times m + m + (m^2+  m) \times (L_d-1).
\end{align*}

\subsection{Combination Layer}
\label{sec:combined}

The combination layer concatenates the outputs from two networks and feed the concatenated vector into a standard logits layer.

The following is the formula for a two-class classification problem: 
\begin{equation}
\label{eq:logits}
	p = \sigma\left([\vecx_{L_1}^T, \vech_{L_2}^T] \vecw_{\text{logits}}  \right),
\end{equation}
where $\vecx_{L_1} \in \mathbb{R}^{d}, \vech_{L_2} \in \mathbb{R}^{m}$ are the outputs from the cross network and deep network, respectively, $\vecw_{\text{logits}} \in \mathbb{R}^{(d + m)}$ is the weight vector for the combination layer, and $\sigma(x) = 1/(1+\exp(-x))$.

The loss function is the log loss along with a regularization term,
\begin{equation}
	\begin{split}
	\text{loss} = &-\frac{1}{N} \sum_{i = 1}^N y_i \log(p_i) + (1 - y_i) \log(1 - p_i) + 
    \lambda \sum_l \|\vecw_l\|^2,
    \end{split}
\end{equation}
where $p_i$'s are the probabilities computed from \autoref{eq:logits}, $y_i$'s are the true labels, $N$ is the total number of inputs, and $\lambda$ is the $L_2$ regularization parameter. 

We jointly train both networks, as this allows each individual network to be aware of the others during the training.


\section{Cross Network Analysis}
\label{sec:analysis_cross_network}
In this section, we analyze the cross network of DCN for the purpose of understanding its effectiveness. We offer three perspectives: polynomial approximation, generalization to FMs, and efficient projection. For simplicity, we assume $\vecb_i = 0$. 

\emph{Notations.} Let the $i$-th element in $\vecw_j$ be $w_{j}^{(i)}$. For multi-index $\vecalpha = [\alpha_1, \cdots, \alpha_d] \in \mathbb{N}^d$ and $\vecx = [x_1, \cdots, x_d] \in \mathbb{R}^d$, we define $|\vecalpha| = \sum_{i=1}^d \alpha_i$.

\emph{Terminology.} The degree of a cross term (monomial) $x_1^{\alpha_1}x_2^{\alpha_2}\cdots x_d^{\alpha_d}$ is defined by $|\vecalpha|$. The degree of a polynomial is defined by the highest degree of its terms.  

\subsection{Polynomial Approximation}
By the Weierstrass approximation theorem \cite{rudin1964principles}, any function under certain smoothness assumption can be approximated by a polynomial to an arbitrary accuracy. Therefore, we analyze the cross network from the perspective of polynomial approximation.  
In particular, the cross network approximates the polynomial class of the same degree in a way that is efficient, expressive and generalizes better to real-world datasets.

We study in detail the approximation of a cross network to the polynomial class of the same degree. Let us denote by $P_n(\vecx)$ the multivariate polynomial class of degree $n$: 
\begin{equation}
P_{n}(\vecx) = \biggl\{\sum_{\vecalpha} w_{\vecalpha} x_1^{\alpha_1}x_2^{\alpha_2}\ldots x_d^{\alpha_d} \mathrel{\bigg|}   0 \le |\vecalpha| \le n, \vecalpha \in \mathbb{N}^d \biggr\}.
\end{equation}
Each polynomial in this class has $O(d^n)$ coefficients. We show that, with only $O(d)$ parameters, the cross network contains all the cross terms occurring in the polynomial of the same degree, with each term's coefficient distinct from each other. 
\begin{theorem}
\label{thm:cross_x0}
	Consider an $l$-layer cross network with the $i+1$-th layer defined as $\vecx_{i+1} =\vecx_{0}\vecx_{i}^T \vecw_{i} +\vecx_{i}$. Let the input to the network be $\vecx_0 = [x_1, x_2, \ldots, x_d]^T$, the output be $g_l(\vecx_0) = \vecx_l^T \vecw_l$, and the parameters be $\vecw_i, \vecb_i \in \mathbb{R}^d$. Then, the multivariate polynomial $g_l(\vecx_0)$ reproduces polynomials in the following class:
    $$\biggl\{\sum_{\vecalpha} c_{\vecalpha}(\vecw_0, \ldots, \vecw_l) x_1^{\alpha_1}x_2^{\alpha_2}\ldots x_d^{\alpha_d} \mathrel{\bigg|}  0 \le |\vecalpha| \le l+1, \vecalpha \in \mathbb{N}^d \biggr\},$$
    where $c_{\vecalpha} = M_{\vecalpha}\sum_{\veci \in B_\vecalpha}\sum_{\vecj \in P_\vecalpha } \prod_{k=1}^{|\vecalpha|}  w_{i_k}^{(j_k)}$, $M_\vecalpha$ is a constant independent of $\vecw_i$'s, $\veci = [i_1, \ldots, i_{|\vecalpha|}]$ and $\vecj = [j_1, \ldots, j_{|\vecalpha|}]$ are multi-indices, $B_{\vecalpha} = \bigl\{\vecy \in \{0,1,\cdots, l\}^{|\vecalpha|} \mathrel{\big|} y_i < y_j \wedge y_{|\vecalpha|} = l\bigr\}$, and $P_\vecalpha$ is the set of all the permutations of  the indices $(\underbrace{1, \cdots, 1}_{\alpha_1 \,\text{times}} \cdots \underbrace{d, \cdots, d}_{\alpha_d\,\text{times}})$. 
\end{theorem}

The proof of \autoref{thm:cross_x0} is in the Appendix.
Let us give an example. Consider the coefficient $c_{\vecalpha}$ for $x_1x_2x_3$ with $\vecalpha = (1,1,1,0,\ldots, 0)$. Up to some constant, when $l=2$, 
$c_\vecalpha = \sum_{i,j,k \in P_\vecalpha} w_0^{(i)}w_1^{(j)}w_2^{(k)}$; when $l=3$, $c_\vecalpha = \sum_{i,j,k \in P_\vecalpha} w_0^{(i)}w_1^{(j)}w_3^{(k)} + w_0^{(i)}w_2^{(j)}w_3^{(k)} + w_1^{(i)}w_2^{(j)}w_3^{(k)}$.

\subsection{Generalization of FMs}
The cross network shares the spirit of parameter sharing as the FM model and further extends it to a deeper structure. 

In a FM model, feature $x_i$ is associated with a weight vector $\vecv_i$, and the weight of cross term $x_ix_j$ is computed by $\langle \vecv_i, \vecv_j\rangle$. In {DCN}, $x_i$ is associated with scalars $\{w_k^{(i)}\}_{k=1}^l$, and the weight of $x_ix_j$ is the multiplications of parameters from the sets $\{w_{k}^{(i)}\}_{k=0}^l$ and $\{w_{k}^{(j)}\}_{k=0}^l$. Both models have each feature learned some parameters independent from other features, and the weight of a cross term is a certain combination of corresponding parameters.  

Parameter sharing not only makes the model more efficient, but also enables the model to generalize to unseen feature interactions and be more robust to noise. For example, take datasets with sparse features.  If two binary features $x_i$ and $x_j$ rarely or never co-occur in the training data, \emph{i.e.}, $x_i \neq 0 \wedge x_j \neq 0$, then the learned weight of $x_ix_j$ would carry no meaningful information for prediction.

 The FM is a shallow structure and is limited to representing cross terms of degree 2. {DCN}, in contrast, is able to construct all the cross terms $x_1^{\alpha_1}x_2^{\alpha_2}\ldots x_d^{\alpha_d}$ with degree $|\vecalpha|$ bounded by some constant determined by layer depth, as claimed in \autoref{thm:cross_x0}. Therefore, the cross network extends the idea of parameter sharing from a single layer to multiple layers and high-degree cross-terms. Note that different from the higher-order FMs, the number of parameters in a cross network only grows linearly with the input dimension.

\subsection{Efficient Projection} 
Each cross layer projects all the pairwise interactions between $\vecx_0$ and $\vecx_l$, in an efficient manner, back to the input's dimension.

	Consider $\tilde \vecx \in \mathbb{R}^{d}$ as the input to a cross layer. The cross layer first implicitly constructs $d^2$ pairwise interactions $x_i\tilde x_j$, and then implicitly projects them back to dimension $d$ in a memory-efficient way. A direct approach, however, comes with a cubic cost.
    
    Our cross layer provides an efficient solution to reduce the cost to linear in dimension $d$. Consider $\vecx_p = \vecx_0 \tilde \vecx^T \vecw$. This is in fact equivalent to 
    \begin{equation}
    	\begin{split}
    	\vecx_p^T = 
        \begin{bmatrix}
        	x_1\tilde x_1 \ldots x_1 \tilde x_d & \ldots & x_d \tilde x_1 \ldots x_d \tilde x_d
        \end{bmatrix}
        \left[\begin{smallmatrix}
        	\begin{smallmatrix} \mid \\ \vecw \\ \mid \end{smallmatrix} & \veczero & \ldots & \veczero \\
            \veczero & \begin{smallmatrix} \mid \\ \vecw \\ \mid \end{smallmatrix} & \ldots & \veczero \\
            \vdots   & \vdots& \ddots & \vdots \\
            \veczero & \veczero & \ldots & \begin{smallmatrix} \mid \\ \vecw \\ \mid \end{smallmatrix}
        \end{smallmatrix}\right]
        \end{split}
    \end{equation}
    
    where the row vector contains all $d^2$ pairwise interactions $x_i\tilde x_j$'s, the projection matrix has a block diagonal structure with $\vecw \in \mathbb{R}^{d}$ being a column vector.

\section{Experimental Results}
\label{sec:experimental_results}
In this section, we evaluate the performance of {DCN} on some popular classification datasets.

\subsection{Criteo Display Ads Data}
The Criteo Display Ads\footnote{https://www.kaggle.com/c/criteo-display-ad-challenge} dataset is for the purpose of predicting ads click-through rate. It has 13 integer features and 26 categorical features where each category has a high cardinality. For this dataset, {\bf an improvement of 0.001 in logloss is considered as practically significant}. When considering a large user base, a small improvement in prediction accuracy can potentially lead to a large increase in a company's revenue.
The data contains 11 GB user logs from a period of 7 days ($\sim$41 million records). We used the data of the first 6 days for training, and randomly split day 7 data into validation and test sets of equal size.

\subsection{Implementation Details} \label{sec:implementation_detail}
{DCN} is implemented on TensorFlow, we briefly discuss some implementation details for training with {DCN}. 

\begin{itemize}[leftmargin=0em]
\item[]	\emph{Data processing and embedding.} Real-valued features are normalized by applying a log transform. For categorical features, we embed the features in dense vectors of dimension $6 \times (\text{category cardinality})^{1/4}.$ Concatenating all embeddings results in a vector of dimension 1026.
\vspace{.1em}
\item[] \emph{Optimization.} We applied mini-batch stochastic optimization with Adam optimizer \cite{kingma2014adam}. The batch size is set at 512. Batch normalization \cite{ioffe2015batch} was applied to the deep network and gradient clip norm was set at 100.
\vspace{.1em}
\item[] \emph{Regularization.} We used early stopping, as we did not find $L_2$ regularization or dropout to be effective.
\vspace{.1em}
\item[]	\emph{Hyperparameters.}
We report results based on a grid search over the number of hidden layers, hidden layer size, initial learning rate and number of cross layers. The number of hidden layers ranged from 2 to 5, with hidden layer sizes from 32 to 1024. For {DCN}, the number of cross layers\footnote{More cross layers did not lead to significant improvement, so we restrict ourselves in a small range for finer tuning.} is from 1 to 6. The initial learning rate\footnote{Experimentally we observe that for the Criteo dataset, a learning rate larger than 0.001 usually degrades the performance.} was tuned from 0.0001 to 0.001 with increments of 0.0001. All experiments applied early stopping at training step 150,000, beyond which overfitting started to occur.
\end{itemize}

\subsection{Models for Comparisons}

We compare {DCN} with five models: the {DCN} model with no cross network ({DNN}), logistic regression (LR), Factorization Machines (FMs), Wide and Deep Model (W\&D), and Deep Crossing (DC).

\begin{itemize}[leftmargin=0em]
\item[] \emph{{DNN}}. The embedding layer, the output layer, and the hyperparameter tuning process are the same as {DCN}. The only change from the {DCN} model was that there are no cross layers.
\vspace{.5em}
\item[] \emph{{LR}}. We used Sibyl \cite{sibyl}---a large-scale machine-learning system for distributed logistic regression. The integer features were discretized on a log scale. The cross features were selected by a sophisticated feature selection tool. All of the single features were used.
\vspace{.5em}
\item[] \emph{FM}. We used an FM-based model with proprietary details.
\vspace{.5em}
\item[] \emph{W\&D}. Different than {DCN}, its wide component takes as input raw sparse features, and relies on exhaustive searching and domain knowledge to select predictive cross features. We skipped the comparison as no good method is known to select cross features. 
\vspace{.5em}
\item[] \emph{DC}. Compared to {DCN}, {DC} does not form explicit cross features. It mainly relies on stacking and residual units to create implicit crossings. We applied the same embedding (stacking) layer as {DCN}, followed by another ReLu layer to generate input to a sequence of residual units. The number of residual units was tuned form 1 to 5, with input dimension and cross dimension from 100 to 1026.
\end{itemize}

\subsection{Model Performance}
In this section, we first list the best performance of different models in logloss, then we compare {DCN} with {DNN} in detail, that is, we investigate further into the effects introduced by the cross network. 

{\bf Performance of different models.}
The best test logloss of different models are listed in \autoref{tab:best_logloss}.
The optimal hyperparameter settings were 2 deep layers of size 1024 and 6 cross layers for the {DCN} model, 5 deep layers of size 1024 for the {DNN}, 5 residual units with input dimension 424 and cross dimension 537 for the {DC}, and 42 cross features for the {LR} model. That the best performance was found with the deepest cross architecture suggests that the higher-order feature interactions from the cross network are valuable. As we can see, {DCN} outperforms all the other models by a large amount. In particular, it outperforms the state-of-art {DNN} model but uses only 40\% of the memory consumed in {DNN}. 

\begin{table}[htpb]
\caption{Best test logloss from different models. ``{DC}" is deep crossing, ``{DNN}" is {DCN} with no cross layer, ``FM'' is Factorization Machine based model, ``LR'' is logistic regression.}
\label{tab:best_logloss}
\begin{center}
\begin{tabular}{c|cccccc}
\toprule
{\bf Model}  &  {DCN} & {DC} & {DNN} & {FM} &  {LR} \\
\midrule
{\bf Logloss} &{\bf 0.4419} & 0.4425 & 0.4428  &0.4464 & 0.4474\\
\bottomrule
\end{tabular}
\end{center}
\end{table}

For the optimal hyperparameter setting of each model, we also report the mean and standard deviation of the test logloss out of 10 independent runs: 
{DCN}: ${\bf 0.4422 \pm 9 \times 10^{-5}}$, {DNN}: $0.4430 \pm 3.7 \times 10^{-4}$, {DC}: $0.4430 \pm 4.3 \times 10^{-4}$. As can be seen, {DCN} consistently outperforms other models by a large amount.

{\bf Comparisons Between {DCN} and {DNN}.}
Considering that the cross network only introduces $O(d)$ extra parameters, we compare {DCN} to its deep network---a traditional {DNN}, and present the experimental results while varying memory budget and loss tolerance. 

In the following, the loss for a certain number of parameters is reported as the best validation loss among all the learning rates and model structures. The number of parameters in the embedding layer was omitted in our calculation as it is identical to both models. 

 \autoref{tab:nparams_vs_logloss} reports the minimal number of parameters needed to achieve a desired logloss threshold. From \autoref{tab:nparams_vs_logloss}, we see that {DCN} is nearly an order of magnitude more memory efficient than a single {DNN}, thanks to the cross network which is able to learn bounded-degree feature interactions more efficiently.

\begin{table}[htpb]
\caption{\#parameters needed to achieve a desired logloss.}
\label{tab:nparams_vs_logloss}
\begin{center}
\begin{tabular}{c|cccc}
\toprule
{\bf Logloss} & 0.4430 & 0.4460 & 0.4470 & 0.4480\\
\midrule
DNN & $3.2 \times 10^6$ & $1.5 \times 10^5$ & $1.5 \times 10^5$ & $7.8 \times 10^4$ \\
DCN & ${\bf 7.9 \times 10^5}$ & ${\bf 7.3 \times 10^4}$ & ${\bf 3.7 \times 10^4}$ & ${\bf 3.7 \times 10^4}$ \\
\bottomrule
\end{tabular}
\end{center}
\end{table}

 \autoref{tab:logloss_vs_nparams} compares performance of the neural models subject to fixed memory budgets. As we can see, {DCN} consistently outperforms {DNN}. In the small-parameter regime, the number of parameters in the cross network is comparable to that in the deep network, and the clear improvement indicates that the cross network is more efficient in learning effective feature interactions. In the large-parameter regime, the DNN closes some of the gap; however, {DCN} still outperforms {DNN} by a large amount, suggesting that it can efficiently learn some types of meaningful feature interactions that even a huge {DNN} model cannot.

\begin{table}[htpb]
\caption{Best logloss achieved with various memory budgets.}
\label{tab:logloss_vs_nparams}
\begin{center}
\begin{tabular}{c|ccccc}
\toprule
{\bf \#Params} & $5\times 10^4$ & $1 \times 10^5$ & $4 \times 10^5$& $1.1 \times 10^6$ & $2.5 \times 10^6$\\
\midrule
DNN & 0.4480 & 0.4471 & 0.4439 & 0.4433 & 0.4431\\
DCN & {\bf 0.4465} & {\bf 0.4453} & {\bf 0.4432} & {\bf 0.4426} & {\bf 0.4423} \\
\bottomrule
\end{tabular}
\end{center}
\end{table}

We analyze {DCN} in finer detail by illustrating the effect from introducing a cross network to a given {DNN} model. We first compare the best performance of {DNN} with that of {DCN} under the same number of layers and layer size, and then for each setting, we show how the validation logloss changes as more cross layers are added. \autoref{tab:change_in_logloss} shows the differences between the {DCN} and {DNN} model in logloss. Under the same experimental setting, the best logloss from the {DCN} model consistently outperforms that from a single {DNN} model of the same structure. That the improvement is consistent for all the hyperparameters has mitigated the randomness effect from the initialization and stochastic optimization. 

\begin{table}[htpb]
\caption{Differences in the validation logloss ($\times 10^{-2}$) between {DCN} and {DNN}.  The {DNN} model is the {DCN} model with the number of cross layers set to 0. Negative values mean that the {DCN} outperforms {DNN}.}
\label{tab:change_in_logloss}
\begin{center}
\begin{tabular}{c|cccccc}
\toprule
\diagbox[width=6.5em]{\bf \#Layers}{\bf \#Nodes}  &    32 & 64 &  128 &   256 &       512 &      1024 \\
\midrule
2&   -0.28& -0.10&  -0.16&   -0.06&   -0.05&   -0.08\\
3&   -0.19& -0.10&  -0.13&   -0.18&   -0.07&   -0.05\\
4&   -0.12& -0.10&  -0.06&   -0.09&   -0.09&   -0.21\\
5&   -0.21& -0.11&  -0.13&   -0.00&   -0.06&   -0.02\\
\bottomrule
\end{tabular}
\end{center}
\end{table}

\begin{figure}[htbp]
  \centering
  \includegraphics[width=2.5in]{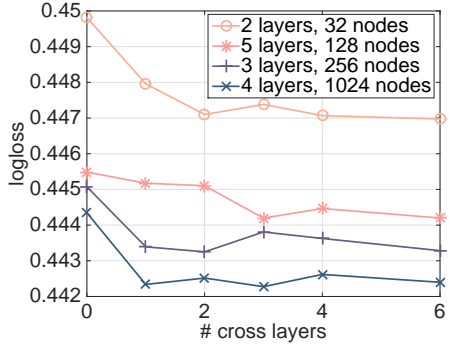}
  \caption{Improvement in the validation logloss with the growth of cross layer depth. The case with 0 cross layers is equivalent to a single {DNN} model. In the legend, ``layers" is hidden layers, ``nodes" is hidden nodes. Different symbols represent different hyperparameters for the deep network.}
  \label{fig:logloss_vs_crosslayers}
\end{figure}

\autoref{fig:logloss_vs_crosslayers} shows the improvement as we increase the number of cross layers on randomly selected settings. For the deep networks in \autoref{fig:logloss_vs_crosslayers}, there is a clear improvement when 1 cross layer is added to the model. As more cross layers are introduced, for some settings the logloss continues to decrease, indicating the introduced cross terms are effective in the prediction; whereas for others the logloss starts to fluctuate and even slightly increase, which indicates the higher-degree feature interactions introduced are not helpful. 

\subsection{Non-CTR datasets}
We show that {DCN} performs well on non-CTR prediction problems. We used the forest covertype (581012 samples and 54 features) and Higgs (11M samples and 28 features) datasets from the UCI repository. The datasets were randomly split into training (90\%) and testing (10\%) set.  A grid search over the hyperparameters was performed. The number of deep layers ranged from 1 to 10 with layer size from 50 to 300. The number of cross layers ranged from 4 to 10. The number of residual units ranged from 1 to 5 with their input dimension and cross dimension from 50 to 300. For {DCN}, the input vector was fed to the cross network directly. 

For the forest covertype data, {DCN} achieved the best test accuracy 0.9740 with the least memory consumption. Both {DNN} and {DC} achieved 0.9737. The optimal hyperparameter settings were 8 cross layers of size 54 and 6 deep layers of size 292 for {DCN}, 7 deep layers of size 292 for {DNN}, and 4 residual units with input dimension 271 and cross dimension 287 for {DC}. 

For the Higgs data, {DCN} achieved the best test logloss 0.4494, whereas {DNN} achieved 0.4506. 
The optimal hyperparameter settings were 4 cross layers of size 28 and 4 deep layers of size 209 for {DCN}, and 10 deep layers of size 196 for {DNN}. {DCN} outperforms {DNN} with half of the memory used in {DNN}.
 
\section{Conclusion and Future Directions}
Identifying effective feature interactions has been the key to the success of many prediction models. Regrettably, the process often requires manual feature crafting and exhaustive searching. {DNN}s are popular for automatic feature learning; however, the features learned are implicit and highly nonlinear, and the network could be unnecessarily large and inefficient in learning certain features. The Deep \& Cross Network proposed in this paper can handle a large set of sparse and dense features, and learns explicit cross features of bounded degree jointly with traditional deep representations. The degree of cross features increases by one at each cross layer. Our experimental results have demonstrated its superiority over the state-of-art algorithms on both sparse and dense datasets, in terms of both model accuracy and memory usage.

We would like to further explore using cross layers as building blocks in other models, enable effective training for deeper cross networks, investigate the efficiency of the cross network in polynomial approximation, and better understand its interaction with deep networks during optimization.


\bibliographystyle{ACM-Reference-Format}
\bibliography{dcn} 


\begin{thebibliography}{00}


\ifx \showCODEN    \undefined \def \showCODEN     #1{\unskip}     \fi
\ifx \showDOI      \undefined \def \showDOI       #1{{\tt DOI:}\penalty0{#1}\ }
  \fi
\ifx \showISBNx    \undefined \def \showISBNx     #1{\unskip}     \fi
\ifx \showISBNxiii \undefined \def \showISBNxiii  #1{\unskip}     \fi
\ifx \showISSN     \undefined \def \showISSN      #1{\unskip}     \fi
\ifx \showLCCN     \undefined \def \showLCCN      #1{\unskip}     \fi
\ifx \shownote     \undefined \def \shownote      #1{#1}          \fi
\ifx \showarticletitle \undefined \def \showarticletitle #1{#1}   \fi
\ifx \showURL      \undefined \def \showURL       #1{#1}          \fi
\providecommand\bibfield[2]{#2}
\providecommand\bibinfo[2]{#2}
\providecommand\natexlab[1]{#1}
\providecommand\showeprint[2][]{arXiv:#2}

\bibitem[\protect\citeauthoryear{Blondel, Fujino, Ueda, and Ishihata}{Blondel
  et~al\mbox{.}}{2016}]%
        {blondel2016higher}
\bibfield{author}{\bibinfo{person}{Mathieu Blondel}, \bibinfo{person}{Akinori
  Fujino}, \bibinfo{person}{Naonori Ueda}, {and} \bibinfo{person}{Masakazu
  Ishihata}.} \bibinfo{year}{2016}\natexlab{}.
\newblock \showarticletitle{Higher-Order Factorization Machines}. In
  \bibinfo{booktitle}{{\em Advances in Neural Information Processing Systems}}.
  \bibinfo{pages}{3351--3359}.
\newblock


\bibitem[\protect\citeauthoryear{Canini}{Canini}{2012}]%
        {sibyl}
\bibfield{author}{\bibinfo{person}{K. Canini}.}
  \bibinfo{year}{2012}\natexlab{}.
\newblock \showarticletitle{Sibyl: A system for large scale supervised machine
  learning.}
\newblock \bibinfo{journal}{{\em Technical Talk\/}} (\bibinfo{year}{2012}).
\newblock


\bibitem[\protect\citeauthoryear{Chapelle, Manavoglu, and Rosales}{Chapelle
  et~al\mbox{.}}{2015}]%
        {chapelle2015simple}
\bibfield{author}{\bibinfo{person}{Olivier Chapelle}, \bibinfo{person}{Eren
  Manavoglu}, {and} \bibinfo{person}{Romer Rosales}.}
  \bibinfo{year}{2015}\natexlab{}.
\newblock \showarticletitle{Simple and scalable response prediction for display
  advertising}.
\newblock \bibinfo{journal}{{\em ACM Transactions on Intelligent Systems and
  Technology (TIST)\/}} \bibinfo{volume}{5}, \bibinfo{number}{4}
  (\bibinfo{year}{2015}), \bibinfo{pages}{61}.
\newblock


\bibitem[\protect\citeauthoryear{Cheng, Koc, Harmsen, Shaked, Chandra, Aradhye,
  Anderson, Corrado, Chai, Ispir, et~al\mbox{.}}{Cheng et~al\mbox{.}}{2016}]%
        {cheng2016wide}
\bibfield{author}{\bibinfo{person}{Heng-Tze Cheng}, \bibinfo{person}{Levent
  Koc}, \bibinfo{person}{Jeremiah Harmsen}, \bibinfo{person}{Tal Shaked},
  \bibinfo{person}{Tushar Chandra}, \bibinfo{person}{Hrishi Aradhye},
  \bibinfo{person}{Glen Anderson}, \bibinfo{person}{Greg Corrado},
  \bibinfo{person}{Wei Chai}, \bibinfo{person}{Mustafa Ispir}, {and}
  \bibinfo{person}{others}.} \bibinfo{year}{2016}\natexlab{}.
\newblock \showarticletitle{Wide \& Deep Learning for Recommender Systems}.
\newblock \bibinfo{journal}{{\em arXiv preprint arXiv:1606.07792\/}}
  (\bibinfo{year}{2016}).
\newblock


\bibitem[\protect\citeauthoryear{He, Zhang, Ren, and Sun}{He
  et~al\mbox{.}}{2015}]%
        {he2015deep}
\bibfield{author}{\bibinfo{person}{Kaiming He}, \bibinfo{person}{Xiangyu
  Zhang}, \bibinfo{person}{Shaoqing Ren}, {and} \bibinfo{person}{Jian Sun}.}
  \bibinfo{year}{2015}\natexlab{}.
\newblock \showarticletitle{Deep residual learning for image recognition}.
\newblock \bibinfo{journal}{{\em arXiv preprint arXiv:1512.03385\/}}
  (\bibinfo{year}{2015}).
\newblock


\bibitem[\protect\citeauthoryear{Ioffe and Szegedy}{Ioffe and Szegedy}{2015}]%
        {ioffe2015batch}
\bibfield{author}{\bibinfo{person}{Sergey Ioffe} {and}
  \bibinfo{person}{Christian Szegedy}.} \bibinfo{year}{2015}\natexlab{}.
\newblock \showarticletitle{Batch normalization: Accelerating deep network
  training by reducing internal covariate shift}.
\newblock \bibinfo{journal}{{\em arXiv preprint arXiv:1502.03167\/}}
  (\bibinfo{year}{2015}).
\newblock


\bibitem[\protect\citeauthoryear{Juan, Lefortier, and Chapelle}{Juan
  et~al\mbox{.}}{2017}]%
        {juan2017field}
\bibfield{author}{\bibinfo{person}{Yuchin Juan}, \bibinfo{person}{Damien
  Lefortier}, {and} \bibinfo{person}{Olivier Chapelle}.}
  \bibinfo{year}{2017}\natexlab{}.
\newblock \showarticletitle{Field-aware factorization machines in a real-world
  online advertising system}. In \bibinfo{booktitle}{{\em Proceedings of the
  26th International Conference on World Wide Web Companion}}. International
  World Wide Web Conferences Steering Committee, \bibinfo{pages}{680--688}.
\newblock


\bibitem[\protect\citeauthoryear{Juan, Zhuang, Chin, and Lin}{Juan
  et~al\mbox{.}}{2016}]%
        {juan2016field}
\bibfield{author}{\bibinfo{person}{Yuchin Juan}, \bibinfo{person}{Yong Zhuang},
  \bibinfo{person}{Wei-Sheng Chin}, {and} \bibinfo{person}{Chih-Jen Lin}.}
  \bibinfo{year}{2016}\natexlab{}.
\newblock \showarticletitle{Field-aware factorization machines for CTR
  prediction}. In \bibinfo{booktitle}{{\em Proceedings of the 10th ACM
  Conference on Recommender Systems}}. ACM, \bibinfo{pages}{43--50}.
\newblock


\bibitem[\protect\citeauthoryear{Kingma and Ba}{Kingma and Ba}{2014}]%
        {kingma2014adam}
\bibfield{author}{\bibinfo{person}{Diederik Kingma} {and}
  \bibinfo{person}{Jimmy Ba}.} \bibinfo{year}{2014}\natexlab{}.
\newblock \showarticletitle{Adam: A method for stochastic optimization}.
\newblock \bibinfo{journal}{{\em arXiv preprint arXiv:1412.6980\/}}
  (\bibinfo{year}{2014}).
\newblock


\bibitem[\protect\citeauthoryear{LeCun, Bengio, and Hinton}{LeCun
  et~al\mbox{.}}{2015}]%
        {lecun2015deep}
\bibfield{author}{\bibinfo{person}{Yann LeCun}, \bibinfo{person}{Yoshua
  Bengio}, {and} \bibinfo{person}{Geoffrey Hinton}.}
  \bibinfo{year}{2015}\natexlab{}.
\newblock \showarticletitle{Deep learning}.
\newblock \bibinfo{journal}{{\em Nature\/}} \bibinfo{volume}{521},
  \bibinfo{number}{7553} (\bibinfo{year}{2015}), \bibinfo{pages}{436--444}.
\newblock


\bibitem[\protect\citeauthoryear{Rendle}{Rendle}{2010}]%
        {rendle2010factorization}
\bibfield{author}{\bibinfo{person}{Steffen Rendle}.}
  \bibinfo{year}{2010}\natexlab{}.
\newblock \showarticletitle{Factorization machines}. In
  \bibinfo{booktitle}{{\em 2010 IEEE International Conference on Data Mining}}.
  IEEE, \bibinfo{pages}{995--1000}.
\newblock


\bibitem[\protect\citeauthoryear{Rendle}{Rendle}{2012}]%
        {rendle:tist2012}
\bibfield{author}{\bibinfo{person}{Steffen Rendle}.}
  \bibinfo{year}{2012}\natexlab{}.
\newblock \showarticletitle{Factorization Machines with {libFM}}.
\newblock \bibinfo{journal}{{\em ACM Trans. Intell. Syst. Technol.\/}}
  \bibinfo{volume}{3}, \bibinfo{number}{3}, Article \bibinfo{articleno}{57}
  (\bibinfo{date}{May} \bibinfo{year}{2012}), \bibinfo{numpages}{22}~pages.
\newblock
\showISSN{2157-6904}


\bibitem[\protect\citeauthoryear{Rudin et~al\mbox{.}}{Rudin
  et~al\mbox{.}}{1964}]%
        {rudin1964principles}
\bibfield{author}{\bibinfo{person}{Walter Rudin} {and}
  \bibinfo{person}{others}.} \bibinfo{year}{1964}\natexlab{}.
\newblock \bibinfo{booktitle}{{\em Principles of mathematical analysis}}.
  Vol.~\bibinfo{volume}{3}.
\newblock \bibinfo{publisher}{McGraw-Hill New York}.
\newblock


\bibitem[\protect\citeauthoryear{Schmidhuber}{Schmidhuber}{2015}]%
        {schmidhuber2015deep}
\bibfield{author}{\bibinfo{person}{J{\"u}rgen Schmidhuber}.}
  \bibinfo{year}{2015}\natexlab{}.
\newblock \showarticletitle{Deep learning in neural networks: An overview}.
\newblock \bibinfo{journal}{{\em Neural networks\/}}  \bibinfo{volume}{61}
  (\bibinfo{year}{2015}), \bibinfo{pages}{85--117}.
\newblock


\bibitem[\protect\citeauthoryear{Shan, Hoens, Jiao, Wang, Yu, and Mao}{Shan
  et~al\mbox{.}}{2016}]%
        {shan2016deep}
\bibfield{author}{\bibinfo{person}{Ying Shan}, \bibinfo{person}{T~Ryan Hoens},
  \bibinfo{person}{Jian Jiao}, \bibinfo{person}{Haijing Wang},
  \bibinfo{person}{Dong Yu}, {and} \bibinfo{person}{JC Mao}.}
  \bibinfo{year}{2016}\natexlab{}.
\newblock \showarticletitle{Deep Crossing: Web-Scale Modeling without Manually
  Crafted Combinatorial Features}. In \bibinfo{booktitle}{{\em Proceedings of
  the 22nd ACM SIGKDD International Conference on Knowledge Discovery and Data
  Mining}}. ACM, \bibinfo{pages}{255--262}.
\newblock


\bibitem[\protect\citeauthoryear{Valiant}{Valiant}{2014}]%
        {valiant2014learning}
\bibfield{author}{\bibinfo{person}{Gregory Valiant}.}
  \bibinfo{year}{2014}\natexlab{}.
\newblock \showarticletitle{Learning polynomials with neural networks}.
\newblock  (\bibinfo{year}{2014}).
\newblock


\bibitem[\protect\citeauthoryear{Veit, Wilber, and Belongie}{Veit
  et~al\mbox{.}}{2016}]%
        {NIPS2016_6556}
\bibfield{author}{\bibinfo{person}{Andreas Veit}, \bibinfo{person}{Michael~J
  Wilber}, {and} \bibinfo{person}{Serge Belongie}.}
  \bibinfo{year}{2016}\natexlab{}.
\newblock \showarticletitle{Residual Networks Behave Like Ensembles of
  Relatively Shallow Networks}.
\newblock In \bibinfo{booktitle}{{\em Advances in Neural Information Processing
  Systems 29}}, \bibfield{editor}{\bibinfo{person}{D.~D. Lee},
  \bibinfo{person}{M.~Sugiyama}, \bibinfo{person}{U.~V. Luxburg},
  \bibinfo{person}{I.~Guyon}, {and} \bibinfo{person}{R.~Garnett}} (Eds.).
  \bibinfo{publisher}{Curran Associates, Inc.}, \bibinfo{pages}{550--558}.
\newblock


\bibitem[\protect\citeauthoryear{Yang and Gittens}{Yang and Gittens}{2015}]%
        {yang2015tensor}
\bibfield{author}{\bibinfo{person}{Jiyan Yang} {and} \bibinfo{person}{Alex
  Gittens}.} \bibinfo{year}{2015}\natexlab{}.
\newblock \showarticletitle{Tensor machines for learning target-specific
  polynomial features}.
\newblock \bibinfo{journal}{{\em arXiv preprint arXiv:1504.01697\/}}
  (\bibinfo{year}{2015}).
\newblock


\end{thebibliography}

\clearpage
{\LARGE \bf Appendix: Proof of Theorem 3.1}
\begin{proof}

{\emph{Notations.}} Let $\veci$ be a multi-index vector of 0's and 1's with its last entry fixed at 1. For multi-index $\vecalpha = [\alpha_1, \cdots, \alpha_d] \in \mathbb{N}^d$ and $\vecx = [x_1, \cdots, x_d]^T$, we define $|\vecalpha| = \sum_{i=1}^d \alpha_i$, and $\vecx^{\vecalpha} = x_1^{\alpha_1}x_2^{\alpha_2}\cdots x_d^{\alpha_d}$.

We first proof by induction that 
\begin{equation}
\label{eq:general_form_cross_x0}
g_l(\vecx_0) = \vecx_l^T \vecw_l = \sum_{p=1}^{l+1} \sum_{\substack{|\veci| = p }} \prod_{j=0}^{l} (\vecx_0^T \vecw_j)^{i_j},
\end{equation}
and then we rewrite the above form to obtain the desired claim. 

\begin{itemize}[leftmargin=0em]
\item[] {\bf Base case.} When $l = 0$, $g_0(\vecx_0) = \vecx_0^T \vecw_0$. Clearly \autoref{eq:general_form_cross_x0} holds.

\item[] {\bf Induction step.} We assume that when $l = k$, 
$$g_k(\vecx_0) = \vecx_k^T \vecw_k = \sum_{p=1}^{k+1} \sum_{\substack{|\veci| = p}} \prod_{j=0}^{k} (\vecx_0^T \vecw_j)^{i_j}.$$ 
When $l = k+1$, 
\begin{equation}
\begin{split}
\vecx_{k+1}^T \vecw_{k+1} = (\vecx_k^T \vecw_k) (\vecx_0^T \vecw_{k+1}) + \vecx_k^T \vecw_{k+1}
\end{split}
\end{equation}
Because $\vecx_k$ only contains $\vecw_0, \ldots, \vecw_{k-1}$, it follows that the formula of $\vecx_{k}^T\vecw_{k+1}$ can be obtained from that of $\vecx_{k}^T\vecw_{k}$ by replacing all the $\vecw_k$'s occurred in $\vecx_{k}^T\vecw_{k}$ to $\vecw_{k+1}$. Then

\begin{equation}
\begin{split}
&\vecx_{k+1}^T \vecw_{k+1} = \\
						& \sum_{p=1}^{k+1} \sum_{\substack{|\veci| = p}}(\vecx_0^T \vecw_{k+1}) \prod_{j=0}^{k} (\vecx_0^T \vecw_j)^{i_j} + 
						\sum_{p=1}^{k+1} \sum_{\substack{|\veci| = p }} (\vecx_0^T \vecw_{k+1})^{i_k}\prod_{j=0}^{k-1} (\vecx_0^T \vecw_j)^{i_j} \\
                        = & \sum_{p=2}^{k+2} \sum_{\substack{|\veci| = p\\ i_k = 1}} \prod_{j=0}^{k+1} (\vecx_0^T \vecw_j)^{i_j} + 
						  \sum_{p=1}^{k+1} \sum_{\substack{|\veci| = p\\ i_{k} = 0}} \prod_{j=0}^{k+1} (\vecx_0^T \vecw_j)^{i_j} \\
                        = &\sum_{p=2}^{k+1} \sum_{\substack{|\veci| = p}} \prod_{j=0}^{k+1} (\vecx_0^T \vecw_j)^{i_j} +  																(\vecx_0^T \vecw_{k+1}) + 
                        \prod_{j=0}^{k+1} (\vecx_0^T \vecw_j) \\
                        = & \sum_{p=1}^{k+2} \sum_{\substack{|\veci| = p}} \prod_{j=0}^{k+1} (\vecx_0^T \vecw_j)^{i_j}.
\end{split}
\end{equation}

The first equality is a result of increasing the size of $\veci$ from $k+1$ to $k+2$. 
The second equality used the fact that the last entry of $\veci$ is always 1 by  definition, and the same was applied to the last equality. 
By induction hypothesis, \autoref{eq:general_form_cross_x0} holds for all $l \in \mathbb{Z}$.
\end{itemize}
Next, we compute $c_\vecalpha(\vecw_0, \cdots, \vecw_l)$, the coefficient of $\vecx^\vecalpha$, by rearranging the terms in \autoref{eq:general_form_cross_x0}. Note that all the different permutations of $\underbrace{x_1 \cdots x_1}_{\alpha_1} \cdots \underbrace{x_d \cdots x_d}_{\alpha_d}$ are in the form of $\vecx^\vecalpha$.  Therefore, $c_\vecalpha$ is the summation of all the weights associated with each permutation occurred in \autoref{eq:general_form_cross_x0}. The weight for permutation $x_{j_1} x_{j_2} \cdots x_{j_p}$ is 
$$\sum_{i_1, \cdots, i_p} w_{i_1}^{(j_1)} w_{i_2}^{(j_2)} \cdots w_{i_p}^{(j_p)},$$ 
where $(i_1, \cdots, i_p)$ belongs to the set of all the corresponding active indices for $|\veci|=p$, specifically, 
$$(i_1, \cdots, i_p) \in B_p =: \bigl\{\vecy \in \{0,1,\cdots, l\}^{p} \mathrel{\big|} y_i < y_j \wedge y_{p} = l\bigr\}.$$ 
Therefore, if we denote $P_\vecalpha$ to be the set of all the permutations of $(\underbrace{1 \cdots 1}_{\alpha_1} \cdots \underbrace{d \cdots d}_{\alpha_d})$, then we arrive at our claim 
\begin{equation}
c_\vecalpha = \sum_{j_1, \cdots, j_p \in P_p} \sum_{i_1, \cdots, i_p \in B_p} \prod_{k=1}^p w_{i_k}^{(j_k)}.
\end{equation}
\end{proof}

\end{document}